\documentclass{article}
\usepackage{spconf,amsmath,graphicx} 

\usepackage{booktabs} 

\usepackage{etoolbox,siunitx}
\robustify\bfseries

\usepackage{tablefootnote} 
\usepackage{multirow}
\usepackage[dvipsnames]{xcolor} 

\usepackage[inline, shortlabels]{enumitem} 
\usepackage{nicefrac}  
\newcommand{\minisection}[1]{\vspace{0.04in} \noindent {\bf #1}\ \ } 

\usepackage{textcomp} 
\usepackage{gensymb}
\usepackage{amssymb} 
\usepackage{bm} 
\usepackage{siunitx} 
\usepackage[export]{adjustbox} 
\usepackage{subfig} 
\usepackage{lipsum}  

\usepackage{diagbox} 

\usepackage{hyperref} 

\definecolor{brickred}{rgb}{0.8, 0.25, 0.33}
\newcommand{\tabred}{\textcolor{brickred}}
\newcommand{\tabyellow}{\textcolor{YellowOrange}}
\newcommand{\tabgreen}{\textcolor{OliveGreen}}

\newcommand{\tabnest}[2]{\begin{tabular}{@{}c@{}}#1\\#2\end{tabular}}  
\newcommand{\tn}{\tabnest}

\makeatletter
\DeclareRobustCommand\onedot{\futurelet\@let@token\@onedot}
\def\@onedot{\ifx\@let@token.\else.\null\fi\xspace}

\makeatother
\usepackage{color}

\usepackage{tikz}  
\newcommand\copyrighttext{
  \footnotesize \textcopyright 2022 IEEE. Personal use of this material is permitted. Permission from IEEE must be obtained for all other uses, in any current or future media, including reprinting/republishing this material for advertising or promotional purposes, creating new collective works, for resale or redistribution to servers or lists, or reuse of any copyrighted component of this work in other works.}
\newcommand\copyrightnoticeprepend{
\begin{tikzpicture}[remember picture,overlay]
\node[anchor=south,yshift=10pt] at (current page.south) {\fbox{\parbox{\dimexpr\textwidth-\fboxsep-\fboxrule\relax}{\copyrighttext}}};
\end{tikzpicture}%
}

\title{MVMO: A MULTI-OBJECT DATASET FOR WIDE BASELINE\\MULTI-VIEW SEMANTIC SEGMENTATION}

\name{Aitor Alvarez-Gila$^{1,2}$\sthanks{This work is part of the projects 3KIA (KK-2020/00049) and BasqNet (KK-2021/00014), funded by the SPRI-Basque Government-ELKARTEK.} \qquad Joost van de Weijer$^{2}$ \qquad Yaxing Wang$^{2}$ \qquad Estibaliz Garrote$^{1}$}
 
\address{$^{1}$ TECNALIA - Basque Research and Technology Alliance (BRTA), Derio, Spain\\
         $^{2}$ Computer Vision Center, Barcelona, Spain}
  
\begin{document}
\copyrightnoticeprepend  

%
\maketitle
\begin{abstract}
We present MVMO (Multi-View, Multi-Object dataset): a synthetic dataset of 116,000 scenes containing randomly placed objects of 10 distinct classes and captured from 25 camera locations in the upper hemisphere.
MVMO comprises photorealistic, path-traced image renders, together with semantic segmentation ground truth for every view.
Unlike existing multi-view datasets, MVMO features wide baselines between cameras and high density of objects, which lead to large disparities, heavy occlusions and view-dependent object appearance.
Single view semantic segmentation is hindered by self and inter-object occlusions that could benefit from additional viewpoints.
Therefore, we expect that MVMO will propel research in multi-view semantic segmentation and cross-view semantic transfer.
We also provide baselines that show that new research is needed in such fields to exploit the complementary information of multi-view setups\footnote{Code and dataset: \href{https://aitorshuffle.github.io/projects/mvmo/}{https://aitorshuffle.github.io/projects/mvmo/}}.
\end{abstract}

\section{Introduction}
\label{sec:intro}

The task of \emph{semantic segmentation}~\cite{longFullyConvolutionalNetworks2015} aims at, given an input image, performing pixel-wise classification over a predefined set of categories.
As in many other dense prediction problems, the end-to-end convolutional neural networks (CNN)-based fully supervised approach to this task has become the \emph{de facto} standard to solve it, leading to robustly performing models~\cite{ronnebergerUNetConvolutionalNetworks2015} at the expense of a large amount of human annotations.
Nevertheless, understanding scenes based on a single 2D input is challenging when applied on
\begin{enumerate*}[label=(\roman*)]
    \item scenes with significant inter-object and self-occlusions that hide class-distinctive features
    \item scenes covering a wide spatial range, where distant objects can show a small apparent size.
\end{enumerate*}

\begin{figure}[ht!]
	\centering
 	\adjincludegraphics[width=1.0\columnwidth]{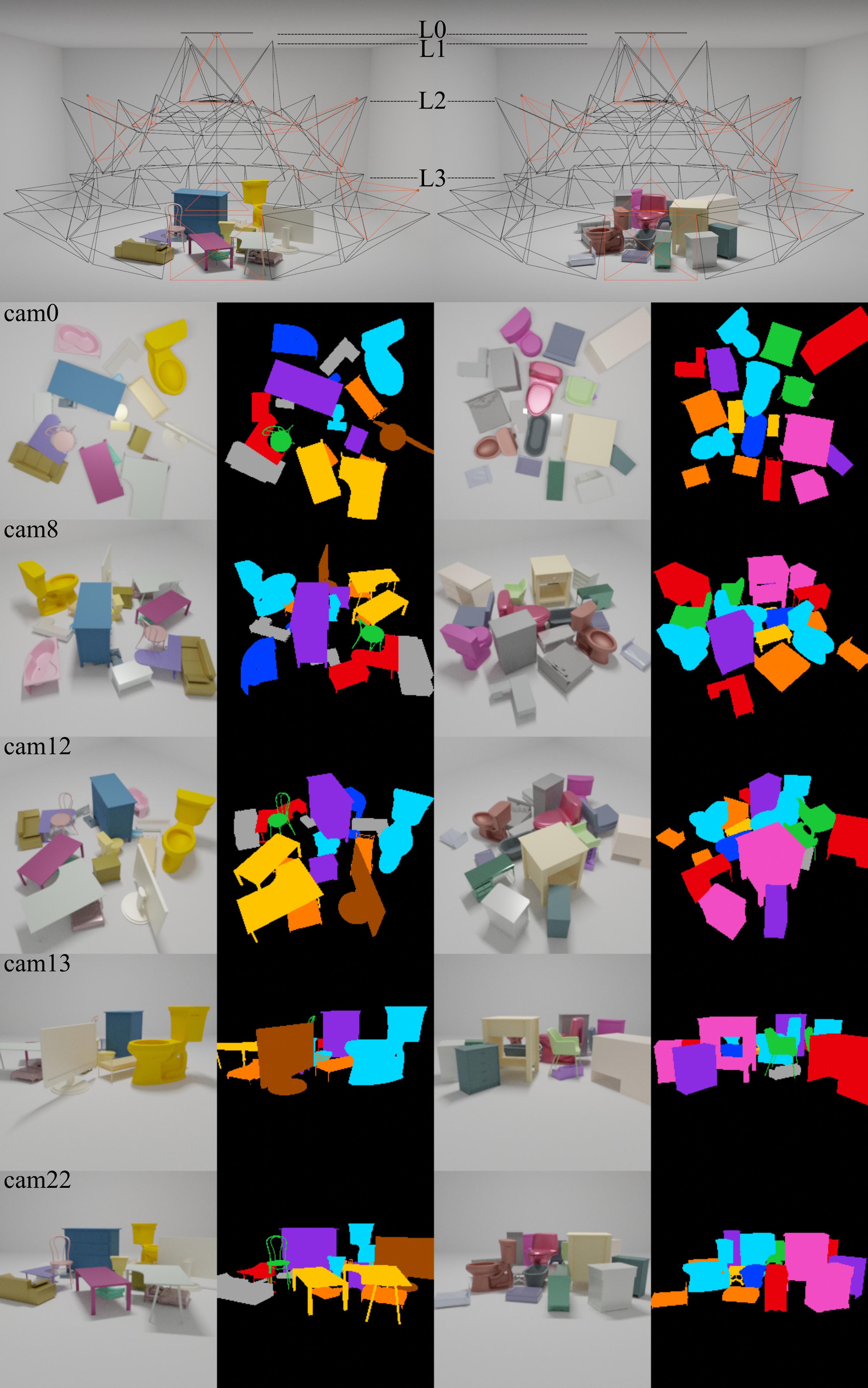}
 	\caption{
 	Top: two scenes from the proposed 116,000 scene MVMO dataset and the 25 equidistributed camera locations. Bottom: rendered views and semantic ground truth for the 5 camera poses (highlighted) used in our experiments.
 	}
	\label{fig_mvmo_samples}
\end{figure}

\begin{table*}[htpb!]  
	\centering
	\resizebox{\textwidth}{!}{
	\sisetup{detect-weight=true,detect-inline-weight=math}
	\begin{tabular}{lccccccc}
		\toprule
		~Dataset~ & 
		~Wide Baseline~ &
		~Object Density~ &
		~Representation~ & 
		~Photorealism~ & 
		~\# Scenes~ & 
		~\# Views~ &
		~\# Classes~\\
		\midrule
		Human3.6M~\cite{ionescuHuman36MLarge2014}       & \tabgreen{Yes} & \tabred{Low (1)} & 2D images & \tabgreen{Real} & 900,000 in 165 sequences & 4 & 24\\ 
        3Dpeople~\cite{pumarola20193dpeople} 		    & \tabgreen{Yes} & \tabred{Low (1)} & 3DM$\rightarrow$2D & \tabgreen{S: High} \tabred{B: Low} & 616,000 in 5,600 sequences & 4 & 8(clothes)/14(body) \\ 
        \midrule
		SYNTHIA~\cite{RosCVPR16}                        & \tabred{No} & N/A & 3DM$\rightarrow$2D & \tabred{Low} & 51,000 in 51 sequences & 8 & 13\\ 
		\midrule
		ScanNet~\cite{daiScanNetRichlyAnnotated3D2017} & \tabyellow{$\star$} & \tabred{Low} & 2D$\rightarrow$3DS & \tabgreen{High} & 1.5k & \tabyellow{$\star$} & 40\\
		House3D~\cite{wuBuildingGeneralizableAgents2018}& \tabyellow{$\star$} & \tabred{Low} & 3DVE & \tabred{Low} & 45.6k & \tabyellow{$\star$} & 80\\ 
		Gibson~\cite{xiaGibsonEnvRealWorld2018} & \tabyellow{$\star$} & \tabred{Low} & 3DVE & \tabgreen{High (IBR/PCR)} & 1.4k & \tabyellow{$\star$} & 40\\  
		CARLA~\cite{dosovitskiy2017carla}               & \tabyellow{$\star$} & \tabyellow{$\star$} & 3DVE & \tabyellow{Mid}-\tabgreen{High} (RT) & \tabyellow{$\star$} & \tabyellow{$\star$} & 12\\
		\midrule
		\textbf{MVMO (ours)}           		            & \tabgreen{Yes} & \tabgreen{High (15-20)} & 3DM$\rightarrow$2D &  \tabgreen{High} (\tabgreen{PT}, \tabyellow{UOM}) & 116k (uncorrelated) & 25 & 11\\
		\bottomrule
	\end{tabular}
	} 
	\caption{\small
	Datasets for multi/cross-view semantic segmentation. The table shows the lack of datasets with wide baseline and high object density addressed by MVMO. \textbf{Object Density}: \#objects/scene. Does not apply to close baseline scenarios. \textbf{Representation}: 2D$\rightarrow$3DS: 3D Surface reconstructed from 2D. 3DVE: 3D Virtual Environment. 3DM$\rightarrow$2D: 3D Model rendered to 2D images. \textbf{Photorealism}: S: Subject. B:Background. IBR: Image-Based Rendering. PCR: Point Cloud Rendering (view synthesis from Point Cloud). RT: Ray-Tracing. PT: Path-Tracing. UOM: Uniform Object Materials. $\star$: Needs to be placed/configured/generated by user; images are not readily available.
	}
	\label{tab_datasets_comparison}
	\vspace{-4mm}
\end{table*}


In this context, we hypothesize that posing data-driven models that exploit multi-view camera setups that provide complementary information over the imaged scenes could be of potential interest for improving the results obtained by single-camera baselines.
However, so far multi-view semantic segmentation has primarily been approached for close-baseline setups~\cite{RosCVPR16} i.e. those where the distance between cameras (and thus, the resulting disparities) are small, whereas solving the aforementioned obstacles requires wide baselines.
Scenarios that could benefit from this approach are frequent in real life, in domains as diverse as industry (e.g. conveyor belts), surveillance, or traffic management.

In this paper, we introduce \textbf{MVMO}, the \textbf{Multi-View Multi-Object dataset}, which addresses the current lack of publicly available large-scale datasets of densely annotated wide-baseline multi-view scenes containing multiple objects.
MVMO is a synthetic, path tracing-based set of 116,000 scenes with per-view semantic segmentation annotations of 10 object categories.
Each scene is observed from $25$ camera locations distributed uniformly in the upper hemisphere (see Fig.~\ref{fig_mvmo_samples}). 
Unlike most existing multi-view image datasets (which are designed to be camera-centric and exhibit very close baselines while sensing their surroundings~\cite{RosCVPR16}), MVMO features wide baselines between many camera pairs as a result of a scene-centric design, and a large amount of objects per scene.
This leads to large disparities, notable occlusions and variable apparent object geometry, size and surface appearances across views.
Therefore, MVMO sets a particularly challenging arrangement that aims at contributing to push research on the fields of multi-view semantic-segmentation and cross-view semantic knowledge transfer.
The experiments presented show that simple baselines fail to be of much help in transferring learned models to novel views, hence suggesting the need for novel research in this direction.

\minisection{Related work.}\label{ssec:related} Our work relates to a number of previous datasets from various research fields, some of which already leverage wide-baseline multi-view datasets in an attempt to improve upon their respective single-view performances:
In multi-view object detection, \cite{roigConditionalRandomFields2011} introduces a multi object detection dataset with bounding box annotations for pedestrians, cars and buses from 6 calibrated cameras.
Advances on multi-view human pose estimation were possible by leveraging various wide baseline datasets over RGB~\cite{kazemiMultiviewBodyPart2013,yaoMONETMultiviewSemiSupervised2019,ionescuHuman36MLarge2014} and depth~\cite{haque2016viewpoint} images of both groups~\cite{jooPanopticStudioMassively2015} and individuals.

The field of \emph{multi-view semantic segmentation} (see Table~\ref{tab_datasets_comparison}) has been addressed from diverse perspectives.
Many early works prior to the irruption of deep learning techniques focused on the binary segmentation of a single static foreground object from a sequence of close-baseline views from a class-agnostic point of view~\cite{campbellAutomatic3DObject2010, leeSilhouetteSegmentationMultiple2011}, often learning sequence-specific models and relying on diverse cues: object-background color distributions, central object fixation or stereo geometry constraints.
More recently, \cite{yaoMultiviewCosegmentationWide2020} used deep self-supervised training to extend the single subject segmentation task to three dynamic scenes in wide-baseline setups.

\emph{Multi-class multi-view semantic segmentation} poses harder challenges and calls for larger datasets. 
Different works leverage the complementary information provided by additional views:
\cite{ladickyJointOptimizationObject2012} extends the Leuven stereo dataset with semantic labels in one of the views to jointly train for segmentation and stereo reconstruction.
A few works focus on \emph{cross-view semantic transfer}, with an unsupervised transfer of the semantic annotations to new label-free views, e.g. ground to aerial views~\cite{zhaiPredictingGroundLevelScene2017} or among distinct vehicle-mounted cameras~\cite{Coors2019THREEDV} in close-baseline footage from~\cite{dosovitskiy2017carla}. 
Both tasks demand datasets comprising two or more 2D RGB views, with annotations in each of them.
SYNTHIA~\cite{RosCVPR16} provides pixel-wise depth and semantic labels for a large synthetic set of scenes captured from a vehicle-mounted 8 RGB camera-rig, thus showing the usual narrow baseline of camera-centric driving setups.
The wide baseline scenario has so far only been tackled by the Human3.6M~\cite{ionescuHuman36MLarge2014} and 3DPeople~\cite{pumarola20193dpeople} datasets.
They both provide body part~\cite{ionescuHuman36MLarge2014,pumarola20193dpeople} or clothing~\cite{pumarola20193dpeople} segmentations, but~\cite{pumarola20193dpeople} has immutable 2D backgrounds, and they are both restricted to single subjects and thus limited in the severity of the occlusions and subject size variation across views.

Several recent papers~\cite{riemenschneiderLearningWhereClassify2014,maMultiviewDeepLearning2017,dai3DMVJoint3DMultiView2018} leverage the spatial consistencies in temporal sequences of RGB or RGB-D images with small relative baselines among them to address semantic segmentation of either 2D images or their reconstructed 3D representations. 
The raw sequences of the NYUv2~\cite{silbermanIndoorSegmentationSupport2012}, Camvid, ETHZ RueMonge 2014~\cite{riemenschneiderLearningWhereClassify2014} or the ScanNet~\cite{daiScanNetRichlyAnnotated3D2017} datasets are commonly used to achieve this.
Furthermore, various large scale 3D virtual or reconstructed environments have been released.
Their relevance comes from the fact that, through significant user intervention, parts of the 3D model and associated labels could be projected back to 2D to synthesise semantically annotated multi-view image sets from arbitrary camera locations with different degrees of realism. 
The House3D~\cite{wuBuildingGeneralizableAgents2018}, Gibson~\cite{xiaGibsonEnvRealWorld2018} and CARLA~\cite{dosovitskiy2017carla} environments are some relevant examples, although only CARLA, being fully virtual, could yield high object densities via its API.
This was shown in~\cite{panCrossviewSemanticSegmentation2019} for close baseline setups, proposing a multi-view semantic fusion scheme from up to 8 input views onto a new virtual zenithal view.

In conclusion, MVMO covers the lack of a standardised large scale photo-realistic multi-view dataset with wide-baselines (and hence, large disparities and relevant occlusions) across cameras and comprising semantic segmentation annotations for multiple objects of distinct classes.

\section{MVMO Dataset construction}
\label{sec:construction}
We use Blender's Python API for procedural 3D scene construction and image rendering, using the ModelNet10 3D object dataset~\cite{wu3DShapeNetsDeep2015} as repository of well-categorized 3D shapes of 10 common object classes. 
We build a basic scene with a grey plane at $z=0$ and a single zenithal rectangular key light, and define a $2.8\times2.8m$ rectangular area for object placement.
All cameras are projective cameras with a focal length of $f=35mm$, oriented to the origin.
The camera locations are determined by sampling the surface of a hemisphere of $r=3m$ regularly so that they are equidistributed~\cite{desernoHowGenerateEquidistributed2004}.
For our set of 25 samples, this yields locations at 4 levels (Fig.~\ref{fig_mvmo_samples}): 1 view at L0 (top, at $z=3.0m$), 3 views at L1 ($z=2.90m$), 9 views at L2 ($z=2.12m$) and 12 views at L3 ($z=0.78m$).

Then, for each scene: 
\begin{enumerate*}[label=(\roman*)]
    \item we randomly select one of the 10 categories of ModelNet10 and
    \item sample one shape from the selected class,
    \item we normalize its scale so that its largest dimension is $1.0m$, then applying a random scale in the $[0.3-0.8]$ range,
    \item we select a random base-color from a set of 9,284 predefined ones and apply a random combination of the \emph{specularity}, \emph{roughness} and \emph{metallic} material modifier properties that -together with other fixed property values- define the Bidirectional Scattering Distribution Function (BSDF) of the materials applied to the whole shape.
    \item we place it on the $z=0$ plane of our base scene, in a random location (within the designated limit area) and angle, checking that the mesh does not intersect with any previously placed object.
    \item Once $15-20$ objects are placed, the scene and fine-detailed ground truth images are rendered with the \textit{Cycles} engine for each of the 25 views at $256\times256$ pixels, producing photo-realistic, unbiased and physically consistent shading, reflectance and material effects, including specularities, and interreflections.
\end{enumerate*}

\begin{figure}[tpb]
\adjincludegraphics[width=0.57\linewidth, valign=t, trim={{0.7cm} {0.7cm} {0.6cm} {0.6cm}}, clip]{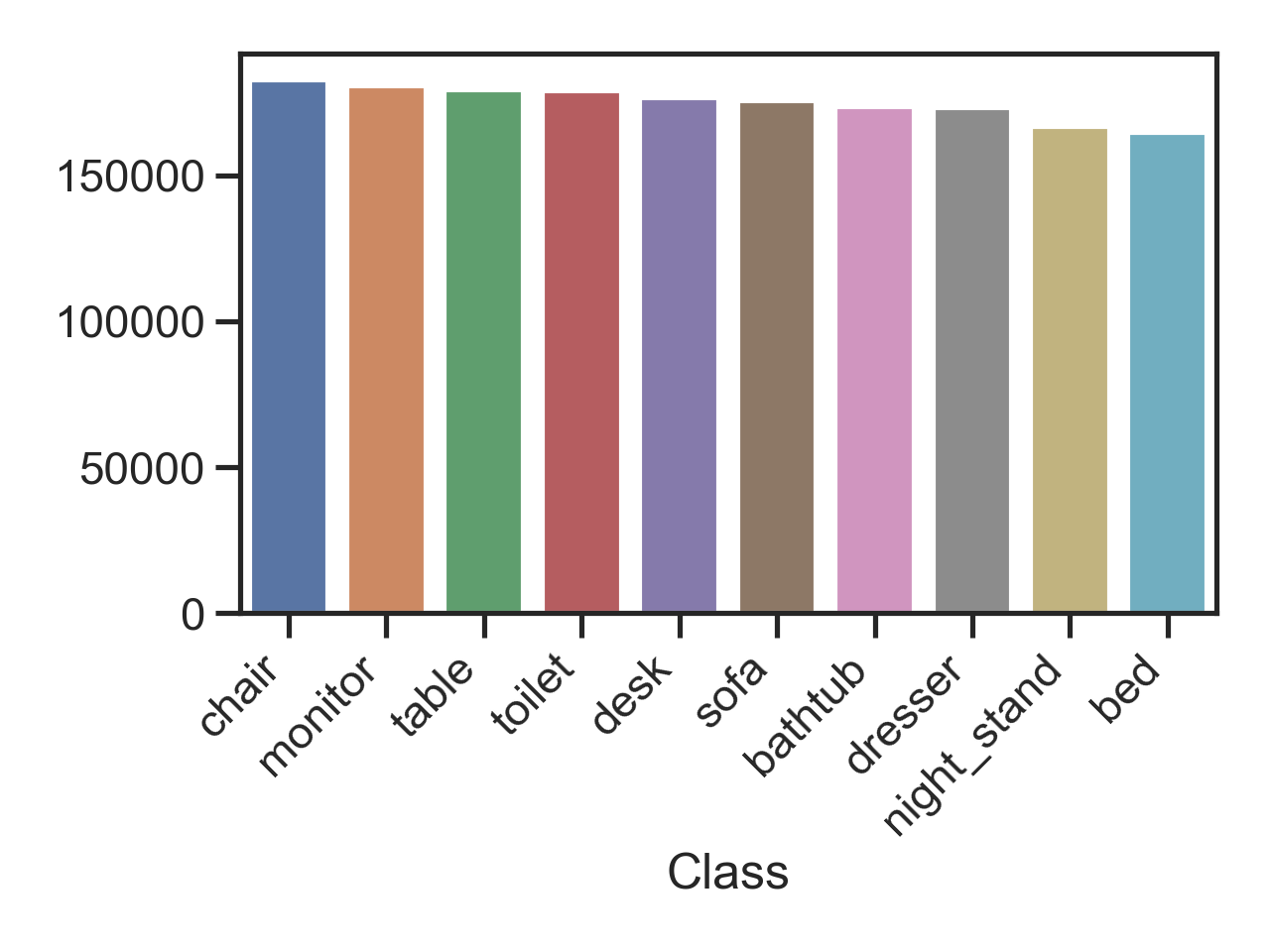}  
\adjincludegraphics[width=0.422\linewidth, height=2.85cm, valign=t, trim={{0.05
cm} {0.6cm} {0.6cm} {0.6cm}}, clip]{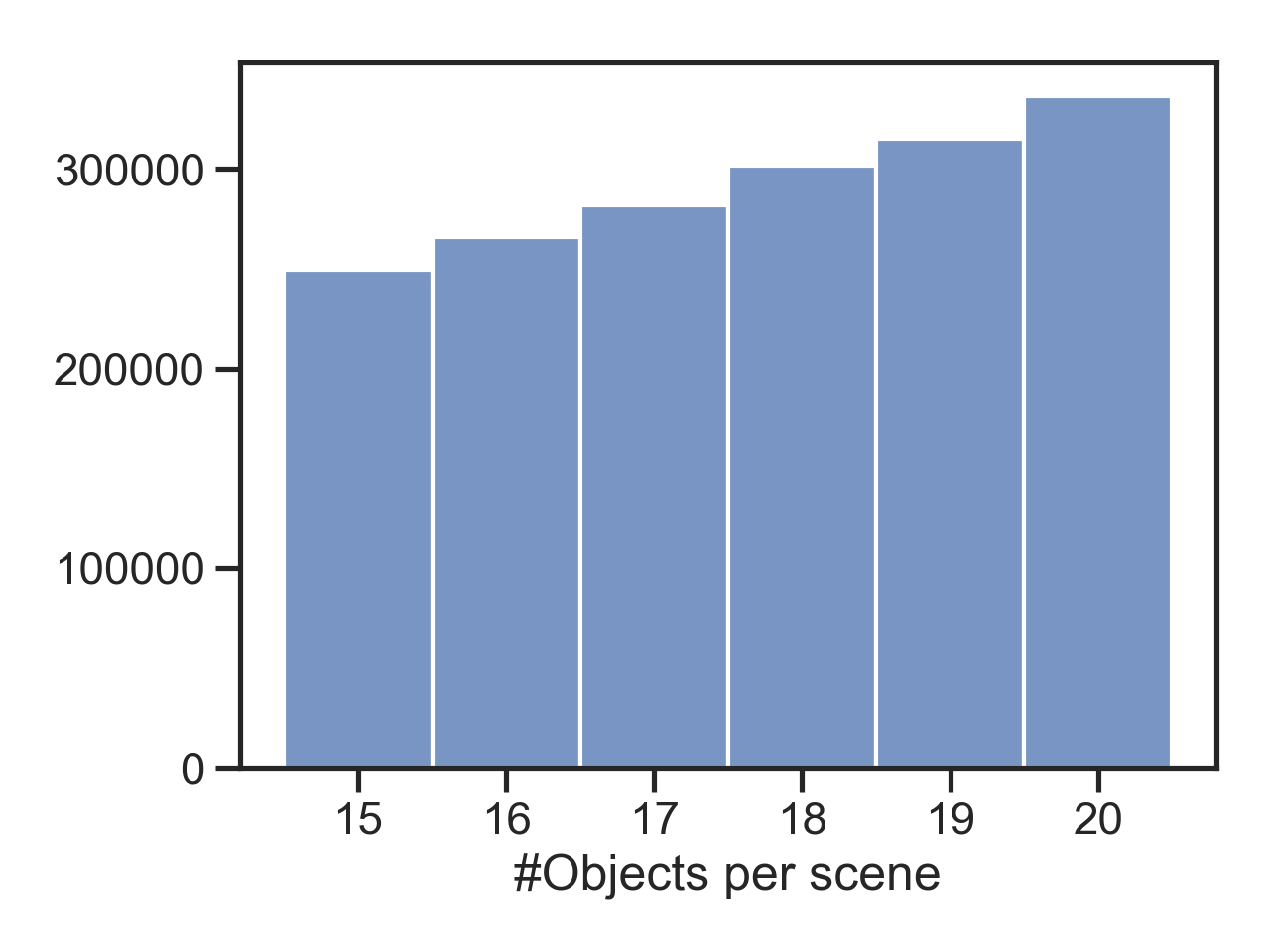}

\caption{Histograms of the train set distributions for (a) Objects per class (total) and (b) Number of objects per scene.}
\label{fig_exploratory_hist}
\vspace{-4mm}

\end{figure}

The 116,000 created scenes (each with 25 views) were then partitioned in a train set (100,000), two validation and two test sets (4,000 each).
The latter are created based on whether the used ModelNet10 shapes were already used for the train set (SO: Same Objects) or come from a held-out set of shapes (OO: Other Objects) from the same categories, which poses a harder problem.
Fig.~\ref{fig_exploratory_hist} shows the resulting distributions of objects per category and scene for the train set.

This proposed wide-baseline multi-object dataset contains many occlusions, making semantic segmentation from a single view difficult.
We think MVMO can facilitate research in multiple directions.
We highlight two of them: 
\begin{enumerate*}[label=(\roman*)]
    \item \emph{Multi-view semantic segmentation}: existing close-baseline datasets have only few occlusions.
    Therefore, the proposed dataset makes for a more interesting setup for multi-view semantic segmentation. 
    \item \emph{Cross-view semantic transfer}: this is an especially exciting research direction which can be performed on MVMO.
    In real-life applications the dense labelling of all views is infeasible.
    Hence we believe that methods need to be designed that can learn to perform multi-view semantic segmentation based on labels from only a single view. 
\end{enumerate*}

\section{Experimental baselines}
\label{sec:experimental}
We run two baseline experiments for the cross-view semantic transfer problem. These experiments are included to show that there is no simple solution to this task and it is indeed an open research problem. To conduct them we select 5 representative views from three distinct levels:  L0.cam0 (zenithal), L2.cam8, L2.cam12, L3.cam.13 and L3.cam.22 (see Fig.~\ref{fig_mvmo_samples}).
In both cases we use a U-Net~\cite{ronnebergerUNetConvolutionalNetworks2015} as our semantic segmentation model, with an Imagenet-pretrained ResNet50 backbone.


\begin{table}[tb]  
	\centering
	\sisetup{detect-weight=true,detect-inline-weight=math}
	\begin{tabular}{llccccc}
		\toprule
		Subset & test\textbackslash train & cam0 & cam8 & cam12 & cam13 & cam22\\
		\midrule
		\multirow{5}{*}{\tn{Other}{objs.}} & L0.cam0 & \textbf{71.12} & 29.09 & 29.61 & 14.28 & 14.88\\
		&L2.cam8 & 24.63 & \textbf{70.21} & \textbf{70.16} & 28.14 & 28.54\\
		&L2.cam12 & 25.14 & 69.09 & 70.05 & 27.73 & 28.29\\
		&L3.cam13 & 12.18 & 31.26 & 31.46 & \textbf{59.18} & 58.72\\
		&L3.cam22 & 12.11 & 30.10 & 30.59 & 58.39 & \textbf{59.41}\\
		\midrule
		\multirow{5}{*}{\tn{Same}{objs.}} & L0.cam0 & \textbf{80.55} & 29.92 & 29.69 & 14.00 & 14.51\\
		&L2.cam8 & 27.11 & \textbf{77.90} & 77.71 & 27.24 & 27.46\\
		&L2.cam12 & 28.01 & 76.87 & \textbf{77.97} & 26.94 & 27.52\\
		&L3.cam13 & 12.90 & 32.16 & 32.29 & \textbf{65.87} & 65.69\\
		&L3.cam22 & 12.76 & 31.00 & 31.68 & 64.84 & \textbf{66.09}\\
		\bottomrule
	\end{tabular}
	\caption{IoU results for direct cross-view semantic transfer. Five models trained on 100\% of the train set (100k scenes).
	}
	\label{tab_exp_direct}
    \vspace{-2mm}
\end{table}



\begin{table}[tb]  
	\centering
	\sisetup{detect-weight=true,detect-inline-weight=math}
    \begin{tabular}{cc cc}
		\toprule
		\multicolumn{2}{c}{L0.cam0$\rightarrow$L2.cam8} & \multicolumn{2}{c}{L2.cam8$\rightarrow$L0.cam0} \\
		\cmidrule(lr){1-2} \cmidrule(lr){3-4} 
		Other objs. & Same objs. & Other objs. & Same objs.\\
		\midrule
		28.72 & 31.29 & 24.35 & 24.84\\
 		\bottomrule
	\end{tabular}
	\caption{IoU results for planar homography-based transfer.}
	\label{tab_exp_homography}
	\vspace{-2mm}
\end{table}

\minisection{Experiment 1. Cross-view semantic transfer via direct testing}
We train an independent model with each of the considered views and directly test them against every other camera's test sets, without any specific adaptation. 
Table~\ref{tab_exp_direct} shows the results in terms of Intersection over Union (IoU):
The diagonals correspond to standard fully supervised single-view setups.
We see that these improve as we adopt a higher perspective of the scene.
As one might expect, direct semantic transfer between cameras placed within the same level (e.g. L2.cam8/L2.cam12) yields a minimal performance drop, on account of the quasi-invariance of the learned representations to horizontal camera pose rotations (the objects were placed in the scene with a random rotation, hence the features observed from both views are similar, except for the non-circular symmetry of the placement area).
However, the performance across views at distinct levels drops drastically, with the most distant levels yielding the highest differences.
Note, finally, the foreseeable performance generalization gap between the OO and SO test subsets that favours the latter.

\begin{figure}[tb]
	\centering
 	
 	 \adjincludegraphics[width=1\columnwidth]{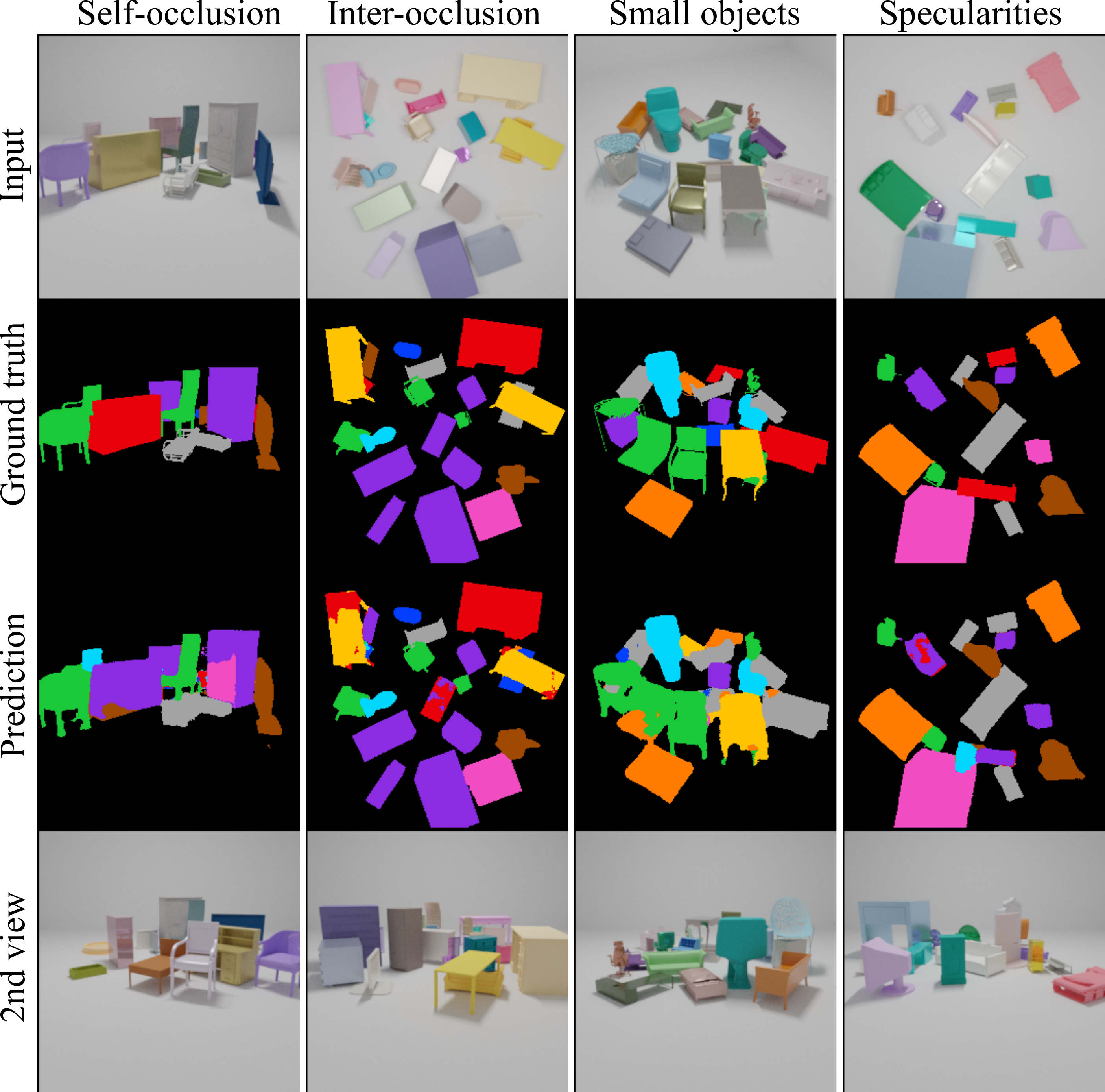} 
 	 	
	\caption{Failure cases from monocular models in Table~\ref{tab_exp_direct}. a) self-occlusion (golden object) b) inter-object occlusion (sofa under the yellow desk) c) small objects (light pink and dark green objects) d) ambiguity from specular inter-reflection (light blue object with reflections of the cyan one). Last row shows a second view that could help solve the ambiguity.}
	\label{fig_exp1_failure_cases}
	\vspace{-2mm}
\end{figure}
Fig.~\ref{fig_exp1_failure_cases} shows some of the most common failure cases for monocular semantic segmentation models:
\begin{enumerate*}[label=(\roman*)]
    \item self-occlusions and 
    \item partial inter-object occlusions that hide relevant features of the object (resulting in ambiguous geometry and appearances),
    \item distant/small objects and, less prominently,
    \item ambiguities induced by appearance variations (e.g. specularities).
\end{enumerate*}
All these cases could benefit from the complementary information provided by the additional, significantly distinct perspectives of a multi-view setup.
Nevertheless, the way of constructively fusing such multiple-view information sources in data-driven models without explicitly addressing a 3D representation of the scene is far from trivial, both in the multi-view and in the cross-view semantic transfer cases. 

\minisection{Experiment 2. Planar homography-based transfer} 
Another baseline to model such geometric relation between views in a cross-view semantic transfer scenario is that of a planar $3\times3$ homography.
This model holds well for quasi-planar scenes or relatively distant objects~\cite{hartleyMultipleViewGeometry2004}.
In this experiment we compute the homography induced by the $z=0$ plane that maps cameras $v_2$ to $v_1$ ($H_{z=0,2\rightarrow1}$) using four point correspondences.
Then, in order to obtain a semantic map estimate from $v_2$ given a model trained on $v_1$ ($f_{v_1\rightarrow ss_1}$), we proceed as follows: 
\begin{enumerate*}[label=(\roman*)]
    \item transform the $v_2$ input to $v_1$ via $H_{z=0,2\rightarrow1}$
    \item feed this to $f_{v_1\rightarrow ss_1}$ so as to obtain a semantic map referenced to $v_1$
    \item transform this back to be referenced to $v_2$ with the inverse homography $H_{z=0,1\rightarrow2}=
    H_{z=0,2\rightarrow1}^{-1}$.
\end{enumerate*}
We test this on two cameras at distinct levels: L0.cam0 and L2.cam8. 
The lack of a significant performance gain in the results (see Table~\ref{tab_exp_homography}) over the direct transfer baseline from Table~\ref{tab_exp_direct} shows that, as expected, the planar homography fails to help for the general, wide-baseline case, in which a good estimate of pixel-wise depth information from every secondary view is needed for unambiguous matching.

The failure of both experimental baselines, along with the fragility of photometric cues in wide baseline scenarios~\cite{yaoMultiviewCosegmentationWide2020}, suggests that exploiting the complementary information given by additional views of the scene in a data-driven multi-view learning setup or transferring the knowledge from trained models across views in unsupervised scenarios will require the development of new theoretical approaches.

\section{Conclusion}
\label{sec:conclusion}
We presented MVMO, a wide baseline multi-view synthetic dataset with semantic segmentation annotations that features a high object density and large amount of occlusions.
We expect MVMO will propel research in multi-view semantic segmentation and cross-view semantic transfer and, likely through domain adaptation, address the current limitations of monocular setups in heavily-occluded real world scenes.


\clearpage  

\bibliographystyle{IEEEbib}
\bibliography{strings,refs}

\end{document}